\documentclass[letterpaper]{article} 
\usepackage{aaai24}  
\usepackage{times}  
\usepackage{helvet}  
\usepackage{courier}  
\usepackage[hyphens]{url}  
\usepackage{graphicx} 
\usepackage{natbib}  
\usepackage{caption} 
\usepackage{url}            
\usepackage{booktabs}       
\usepackage{amsfonts}       
\usepackage{nicefrac}       
\usepackage{microtype}      
\usepackage{xcolor}         
\usepackage{times}
\usepackage{soul}
\usepackage{graphicx}
\usepackage{amsmath}
\usepackage{amsthm}
\usepackage{amssymb}
\usepackage{amsfonts}
\usepackage[ruled,vlined]{algorithm2e}
\usepackage[tableposition=top]{caption}
\usepackage{multirow}
\usepackage{makecell}
\usepackage{booktabs}
\usepackage{enumitem}
\usepackage{footmisc}
\usepackage{subfigure}
\usepackage{appendix}
\usepackage{tabularx}
\usepackage{multirow}
\usepackage{multibib}

\usepackage{inconsolata}
\newcommand\ours{\textsc{TReC}}
\newcommand\ourswb{\textsc{TReC} }
\newtheorem{Degradation}{Definition}

\title{Text Diffusion with Reinforced Conditioning}
\author {
    Yuxuan Liu\textsuperscript{\rm 1},
    Tianchi Yang\textsuperscript{\rm 2},
    Shaohan Huang\textsuperscript{\rm 2},
    Zihan Zhang\textsuperscript{\rm 2},
    Haizhen Huang\textsuperscript{\rm 2}\\
    Furu Wei\textsuperscript{\rm 2},
    Weiwei Deng\textsuperscript{\rm 2},
    Feng Sun\textsuperscript{\rm 2},
    Qi Zhang\textsuperscript{\rm 2}
}
\affiliations {
    \textsuperscript{\rm 1}Peking University \\
    \textsuperscript{\rm 2}Microsoft Corporation \\
    yx.liu@stu.pku.edu.cn
}

\begin{document}
\maketitle

\begin{abstract}
Diffusion models have demonstrated exceptional capability in generating high-quality images, videos, and audio. Due to their adaptiveness in iterative refinement, they provide a strong potential for achieving better non-autoregressive sequence generation. However, existing text diffusion models still fall short in their performance due to a challenge in handling the discreteness of language. This paper thoroughly analyzes text diffusion models and uncovers two significant limitations: degradation of self-conditioning during training and misalignment between training and sampling. Motivated by our findings, we propose a novel \textbf{\underline{T}}ext Diffusion model called \ours, which mitigates the degradation with \textbf{\underline{Re}}inforced \textbf{\underline{C}}onditioning and the misalignment by Time-Aware Variance Scaling. Our extensive experiments demonstrate the competitiveness of \ourswb against autoregressive, non-autoregressive, and diffusion baselines. Moreover, qualitative analysis shows its advanced ability to fully utilize the diffusion process in refining samples.
\end{abstract}

\section{Introduction}
Diffusion models \cite{ho2020denoising, DBLP:conf/iclr/SongME21} are the de facto state-of-the-art generative models in the field of vision \cite{rombach2022high, ho2022imagen} and audio \cite{kongdiffwave, liu2022diffsinger} given their promising capability in generating high-quality samples. 
However, due to the discrete nature of language modality, it is non-trivial to extend diffusion to the field of natural language generation (NLG), and how to empower NLG with diffusion models is becoming a rapidly emerging research area. 

On this front, \citet{austin2021structureddiff} and \citet{hoogeboom2021argmax} design a discrete diffusion process based on categorical distributions, while \citet{he2022diffusionbert} explored diffusion with state absorption (i.e., mask tokens as noise injection). \citet{li2022diffusion} first proposed to directly remedy the discrete nature by mapping words onto a continuous embedding space. However, the above studies only achieved unconditional or coarse-grained control of sequence generation, whose empirical applications are limited. 

Consequently, subsequent works mainly focus on conditional generation, which is a more universally applicable scenario in NLG. Later improvements in the conditioning strategies are mainly categorized three-fold. The first line includes conditioning on controlling attributes, like topics or sentiments \cite{lovelace2022latent, liu2022composable, li2023diffusion}. The second line applies diffusion models to text-to-text generation, i.e., conditioning on input sequences \cite{gong2022diffuseq, yuan2022seqdiffuseq, gao2022difformer, ye2023dinoiser}. This yields more applicable tasks like machine translation or paraphrasing, which are considered more challenging \cite{li2023diffusion}. The third line study conditioning on predictions of previous steps, namely self-conditioning \cite{chen2022analog} to boost model performance.

In this paper, we start by taking a thorough analysis of the vanilla self-conditioning approach and observe it suffers from degradation - marginalizing the diffusion latent. Hampered by such degradation, sampling with self-conditioning heavily depends on the quality of the first step (from pure Gaussian) and fails to fully utilize the diffusion process. Besides, by analyzing current sampling methods in text diffusion models, we discover and study the misalignment issue, bringing out insights in designing a better variance schedule.

Motivated by our findings, we propose \ours, a novel approach that empower \underline{T}ext Diffusion models with \underline{Re}inforced \underline{C}onditioning. Specifically, we develop a novel reinforced self-conditioning that mitigates the degradation by directly motivating quality improvements from self-conditions with reward signals. Furthermore, we propose time-aware variance scaling that facilitates training of diffusion. We conduct a series of experiments on various tasks of NLG, including machine translation, paraphrasing, and question generation. Results show that composing operators within our method manages to generate high-quality sequences, outperforming a series of autoregressive, non-autoregressive, and diffusion baselines. Detailed analysis demonstrates the effectiveness of \ourswb in mitigating degradation of self-conditioning with reward signals, as well as leveraging the diffusion process to iteratively refine its output.

\section{Preliminaries}
\subsection{Denoising Diffusion Probablistic Models}
Denoising diffusion probabilistic models \cite{sohl2015deep, ho2020denoising} learn a series of state transitions from prior data distribution $z_0 \sim q(x)$ to pure Gaussian $z_T \sim \mathcal{N}(0, \mathbf{I})$ through forward and reverse diffusion process. Each forward diffusion step $t \in [1,2,...,T]$ is a Markov process: $q(z_t|z_{t-1})=\mathcal{N}(z_t; \sqrt{1-\beta_t}z_{t-1}, \beta_t \mathbf{I})$, where $\beta_t$ is a schedule for variance scale added at each forward step. Using the superposition property of the Gaussian distribution, we obtain the following closed form for sampling $z_t$ from $z_0$:
\begin{equation}
    q(z_t|z_0)=\mathcal{N}(z_t; \sqrt{1-\bar{\beta}_t}z_{t-1}, \bar{\beta}_t \mathbf{I}),
\end{equation}
where $\bar{\beta}_t:=1-\prod_{i=0}^t\left(1-\beta_i\right)$. In the reverse diffusion process, we learn a denoising function: $p_\theta(z_{t-1}|z_t)=\mathcal{N}(z_{t-1} ; \mu_\theta(z_t, t), \Sigma_\theta(z_t, t))$, where $\mu_\theta$ and $\Sigma_\theta$ denote model's prediction on mean and variance for $z_{t-1}$, respectively. With the reverse process, we could reconstruct $z_0$ by gradually denoising $z_T$ following the trajectory $z_T \to z_{T-1} \to ... \to z_0$. To parameterize the model, we define $\mu_\theta(z_t, t)$ as $\mu(z_{t-1}|z_t, \hat{z_0})$  and predict $\hat{z_0}$ via a neural network: $\hat{z_0} = f_\theta(z_t,t)$. Then we could train the diffusion model through minimizing the prediction error \cite{ho2020denoising}: 
\begin{equation}
    \mathcal{L}_{\text{Diffusion}}(\widehat{z_0}) = \mathbb{E}_{z_0,t}\left[\Vert \hat{z_0}-z_0\Vert^2 \right].
\end{equation}

\subsection{Self-Conditioning}
\label{self-cond}
First proposed in Analog Bits \cite{chen2022analog}, self-conditioning has shown to be an effective method in training denoising diffusion probabilistic models \cite{gao2022difformer, yuan2022seqdiffuseq}. 
Self-conditioning slightly alters the denoising function from $f_\theta (z_t,t)$ to $f_\theta (z_t,\widehat{z_{0}},t)$, to leverage $z_0$ prediction from the previous step. During training, self-conditioning is taken at a certain probability (e.g., 50\%), otherwise the vanilla denoising function $f_\theta (z_t,0,t)$ is trained (setting $\widehat{z}_{0}=\textbf{0}$). At one training step $t \sim U(0,T)$, we first obtain an initial prediction $\widehat{z_{0}}=f_\theta (z_t,0,t)$ , then predict $z_0$ again by feeding the concatenation of $z_t$ and $\widehat{z_{0}}$ into model (i.e., $z_0^{SC}=f_\theta (z_t,\widehat{z_{0}},t))$. Since we only back propagate on $z_{0}^{SC}$, such method could be employed with only a small cost increase during training \cite{chen2022analog}, and that is negligible during sampling.

\subsection{Continuous Diffusion for Text Generation}

Continuous text diffusion models map discrete sequences onto a continuous space (e.g., word vector space) and diffuse over this space \cite{li2022diffusion, gong2022diffuseq, yuan2022seqdiffuseq, gao2022difformer, dieleman2022continuous}. To bring the optimization objective, we could regard diffusion models as variational auto-encoders, and minimizing the evidence lower bound (ELBO) of $\log p_\theta(y)$ \cite{vahdat2021score, wehenkeldiffusion} theoretically as:
\begin{equation}
    \label{loss}
    \mathcal{L}(y) = \mathbb{E}_{y, z_0 \sim q(y)}\left[\mathcal{L}_{\text{Diffusion}}(\widehat{z_0})-\log p_\theta(y | z_0)\right],
\end{equation}
where $y$ is the target sequence. On estimating $\log p_\theta(y)$, \citet{li2022diffusion} first propose to sample $z_0$ from noisy word embedding of $y$: $\mathcal{N}(Emb(y), \beta_0 \textbf{I})$, and address the reconstruction of $y$ with $\log p_\theta(y|z_0)$. \citet{gao2022difformer} found this trivial as the gap between noisy start $z_0$ and $Emb(y)$ is relatively small, and propose to train by directly reconstructing $y$ from model's output, i.e., $\log p_\theta(y|\widehat{z_0})$. In extension to conditioned sequence generation, current approaches alter denoising function $f_\theta(z_t,t)$ by adding source sequence $x$ or controlling attributes $a$ to conditions, i.e.,  $f_\theta(z_t,x,t)$ and $f_\theta(z_t,a,t)$.

\section{Pitfalls of Status Quo}
\subsection{Degradation of Self-Conditioning During Training} \label{deg_ch}

In this section, we recognize and analyze the degradation of self-conditioning during training process of continuous diffusion language models. 
As elaborated in the previous chapter, self-conditioning is designed to utilize near-accurate prediction $\hat{z}_0$ to provide an additional conditional guidance and act as a hint to the denoising function for better denoising. By adding self-condition, we motivate the model to perform better denoising by providing additional information.

\begin{figure}[!t]
  \centering
   \begin{minipage}{\linewidth}
    \subfigure[\label{1a}]{\includegraphics[width=\linewidth]{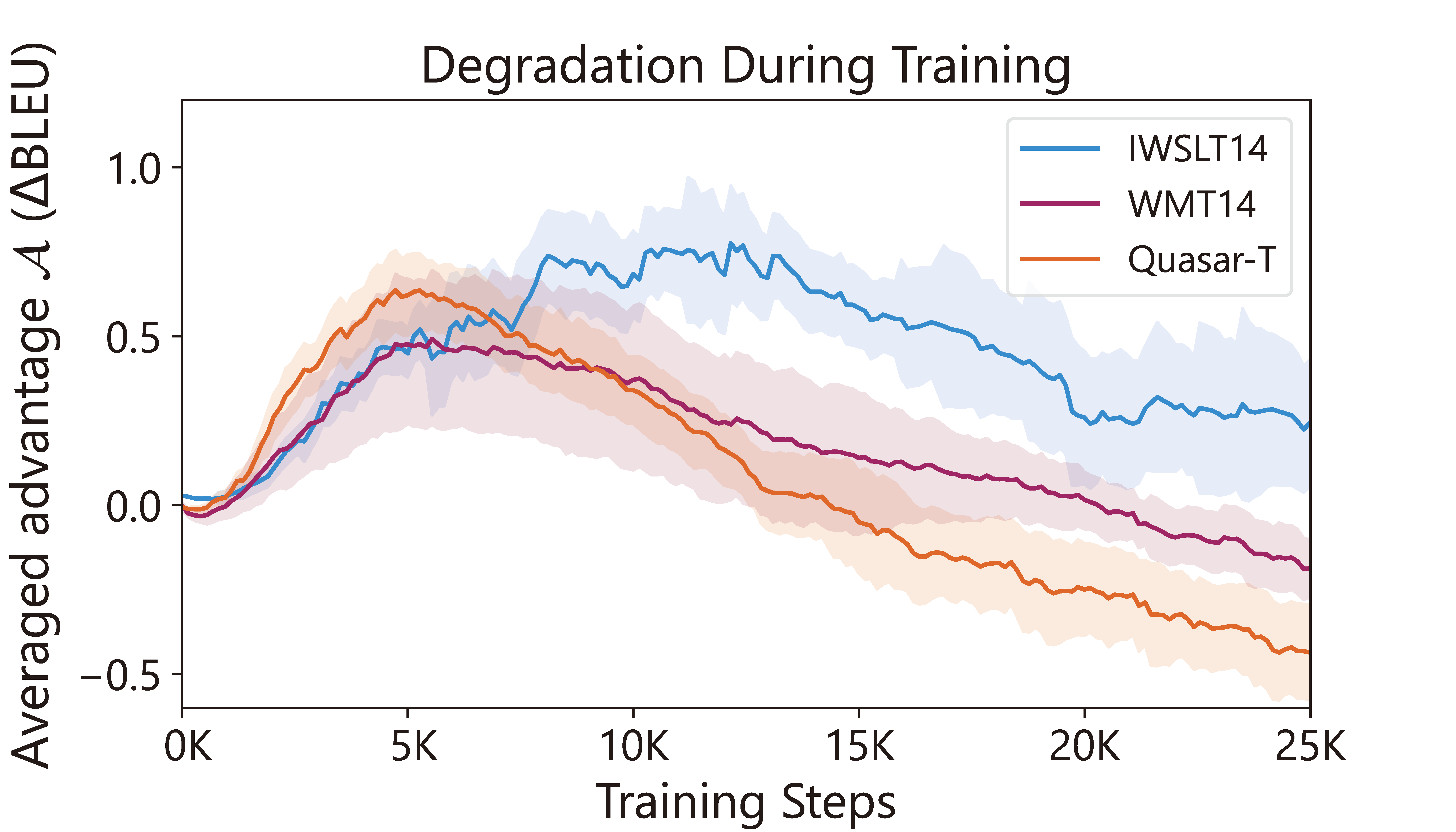}}
    \end{minipage}\
    \begin{minipage}{\linewidth}
    \subfigure[\label{1b}]{\includegraphics[width=\linewidth]{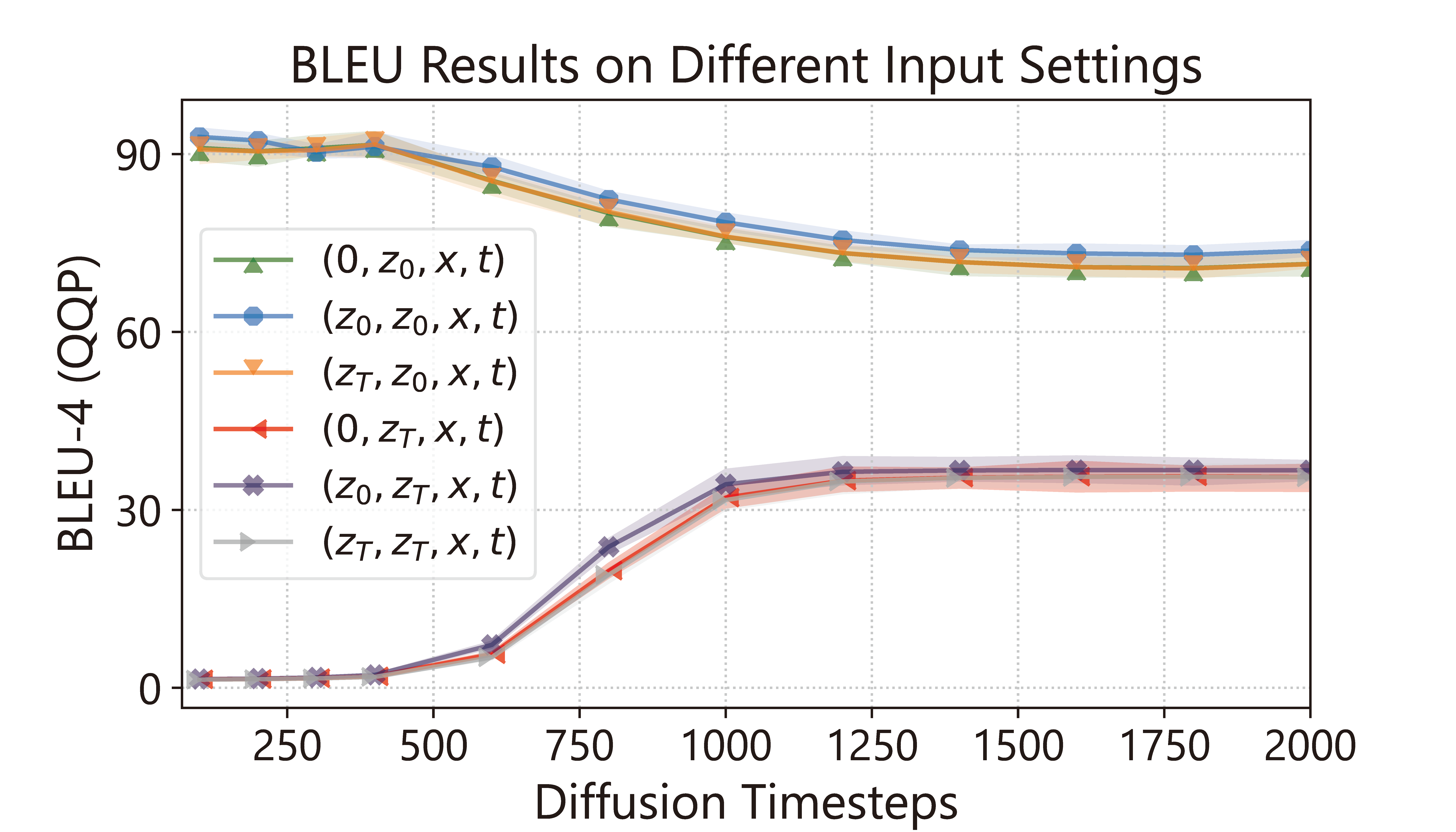}}
    \end{minipage}

  \caption{Degradation of Self-Conditioning. (a) Quality advantage ($\Delta$BLEU) from self-conditioning on valid set during training, which first increases and then decreases. (b) BLEU scores based on inputs $(\forall,z_0,x,t)$ and $(\forall,z_T,x,t)$ constructed from validation samples. This further validates that the model is extremely sensitive to $\hat{z}_0$ (\textit{the first term, previous prediction}) and insensitive to the $z_t$ (\textit{the second term, noised latent}) to be denoised.}
  \label{degrade}
\end{figure}

However, such desired improvements on denoising are not aligned with the
training process, and thus not ensured during sampling. Start by
recalling the training objective
$\mathcal{L}=\mathbb{E}_{z,t} ( ||z_{0}^{SC} - z_{0}|| - \log p(y|z_{0}^{SC}) )$
and denoising step
${z_{0}^{SC} = f}_{\theta}(z_{t},\widehat{z}_{0},x,t)$. When the model
is mostly converged, it can provide near accurate $\hat{z}_{0}$
predictions. 
Even if the self-condition step fails to further optimize $z^{SC}_0$ over $\hat{z}_0$, the total training objective could still converge due to the improving accurateness of $\hat{z}_0$ predictions. Therefore, the self-condition denoising step $f_\theta(z_t, \hat{z}_0 ,x,t)$ could easily achieve a low loss by simply copying $\hat{z}_0$ as its output, as reconstruction from  $\hat{z}_0$ to $z^{SC}_0$ becomes substantially easier when $\hat{z}_0$'s quality increases progressively. To this end, there would be a great tendency for $\pi^{SC}_\theta$ to \textit{marginalize or even ignore} $z_t$, which makes self-conditioned training trivial. We define this phenomenon the \textbf{degradation of
self-conditioning}. 

\begin{Degradation}[Degradation of Self-Condition]
    Denote \(\hat{z}_{0} = f_{\theta}\left( z_{t},0,x,t \right)\) the initial prediction of denoising target \(z_{0}\) without self-conditioning; \(t\) and \(x\) the diffusion step and input condition, respectively. A denoising function $f_{\theta}\left( z_{t},\widehat{z}_{0},x,t \right)$ is \textbf{degraded} if it marginalizes the noised latent term \(z_{t}\). 
\end{Degradation}

To consolidate this analysis, we provide two experimental
observations, as shown in Figure \ref{degrade}. To track the denoising quality during
training phase, we evaluate the quality of $\widehat{z}_{0}$ and
$z_{0}^{SC}$ with a tractable metric (i.e., BLEU), and calculate the
quality improvements $\mathcal{A}$ of self-conditioned denoising over the
initial prediction by
\begin{equation}
\label{adv_estim}
    \mathcal{A} = (BLEU(\hat{y}|z_{0}^{SC},y) -BLEU(\hat{y}|\widehat{z}_{0},y))
\end{equation}
during training. As illustrated in Figure \ref{1a}, the quality advantages first rises then decreases, indicating such degradation do occur during training.
In Figure \ref{1b}, we further assure such degradation by feeding six diverse input combinations of $(z_t, \hat{z}_0)$. As illustrated in the Figure \ref{1b}, performance curves with same $\hat{z}_0$: $z_0$ (ground truth) or $z_T$ (pure Gaussian) \textit{highly overlaps}, showing that given the same $\hat{z}_0$, information provided in $z_t$ provide merely insignificant impact (as there aren't significant difference within group $(0,z_0),(z_0,z_0),(z_T,z_0)$ or $(0,z_T),(z_0,z_T),(z_T,z_T)$). This phenomenon indicates that outputs are heavily conditioned on last-step predictions $\hat{z}_0$, but mostly independent of noised latent $z_t$, which should have been focused instead. Such degradation trivializes the diffusion process, which obviously contradicts the design goals of self-conditioning and diffusion.

\subsection{Misalignment With Training During Sampling}

Sampling is critical in obtaining high-quality outputs for text diffusion models. From NLP perspective, \citet{li2022diffusion} propose rounding trick to match each predicted embedding to its nearest neighbor during sampling to prevent diffusion on non-vocabulary. However, such KNN is time-heavy, and its loss $\mathcal{L}_{round}$ leads to unstable training. From the diffusion side, latest work include asymmetric time intervals \cite{chen2022analog} and noise factor \cite{gao2022difformer}.
Specifically, the former alter the denoising function with small time gap (i.e, from $f_\theta(z_t,t)$ to $f_\theta(z_t,t+\Delta)$), while the latter propose to train with a higher variance prior $\mathcal{N}(0, F^2\mathbf{I}),F \geq 1$, then sample with a smaller one, i.e., $F=1$. 

However, despite their practical gains, they are proposed from a pure empirical perspective without supporting theories, and the in-depth explanations beyond their effects remain under-explored. In this section, we study the misalignment with training during sampling, and derive that existing works \cite{chen2022analog, gao2022difformer} are complementary in terms of mitigating such misalignment, and in preventing such phenomenon brings us clear insights to designing a better sampling regime. 

\begin{Degradation}[Misalignment During Sampling]
    Given data sample $(x,y)$ (e.g., paired sequences), sampling step $t$, and diffusion latent $z_{t+1}$ from the previous step of reverse diffusion. We define $z_{t+1}$ is misaligned with training during sampling, if it becomes a small probability event under the distribution $z_{t+1} \sim \mathcal{N}(z_0; \sqrt{1-\bar{\beta_{t}}}z_0, \bar{\beta_t}^2\mathbf{I})$.
\end{Degradation}

\paragraph{Study on Misalignment During Sampling} Consider a sampling step of diffusion process at given time-step \(t\), in which we sample \(z_{t - 1}\) based on
$$\widehat{z}_{t - 1} \sim q\left( \widehat{z}_{t - 1} \middle| \widehat{z}_{t},\widehat{z}_{0} \right)p_{\theta}\left( \widehat{z}_{0} \middle| \widehat{z}_{t},x,t \right).$$
$q$ denotes the DDIM \cite{DBLP:conf/iclr/SongME21} sampler
$q\left( \hat{z}_{t - 1} \middle| \widehat{z_{t}},\widehat{z}_{0} \right)\mathcal{= N(}\sqrt{1 - \bar{\beta}_{t}^{rev}}\widehat{z}_{0},\ \bar{\beta}_{t}^{rev}\widetilde{\epsilon}_{t})$, and $p_\theta$ denotes the denoising model.
Afterwards, variance \(\epsilon_{t}\) added during forward process \(z_{t}\sim q(z_{t}|z_{0},t)\) is estimated according to 
\begin{equation}
    \widetilde{\epsilon}_{t} = (z_{t} - \sqrt{1 - \bar{\beta}_{t}^{train}}\widehat{z}_{0}) \bigg/ \sqrt{\bar{\beta}_{t}^{train}}.
\end{equation}

The next latent
\(z_{t - 1}\) is deterministicly sampled following
\begin{equation}
    \label{zthat}
    \widehat{z}_{t - 1} = \sqrt{1 - \bar{\beta}_{t-1}^{rev}}\widehat{z}_{0} + \sqrt{\bar{\beta}_{t-1}^{rev}}\widetilde{\epsilon}_{t}.
\end{equation}

Now consider the forward process on \((x,y,t-1)\), we have
\begin{equation}
    z_{t-1} = \sqrt{1 - \bar{\beta}_{t-1}^{train}}z_{0} + \sqrt{\bar{\beta}_{t-1}^{train}}\epsilon_{t-1},
\end{equation}

where \(\epsilon_{t-1}\ \sim \mathcal{N}(0,\mathbf{I})\). During inference, there exists non-negligible prediction error, given that we couldn't reach exact accuracy during inference. Denote $\sigma_{t-1}$ the reconstruction error, we could rewrite Eq.\eqref{zthat} into the following form:
\begin{equation}
    \label{misalignment_prevention}
    \widehat{z}_{t - 1} = \sqrt{1 - \bar{\beta}_{t-1}^{rev}}z_0 + (\sqrt{1 - \bar{\beta}_{t-1}^{rev}}\sigma_{t-1} + \sqrt{\bar{\beta}_{t-1}^{rev}}\widetilde{\epsilon}_{t}).
\end{equation}

Given that we could not achieve 100\% inference accuracy, such predicted error would be addressed as part of added noise in training. We could thus improve sampling by preventing misalignment: the input $\hat{z}_t$, given the non-negligible prediction error $\sigma$, should not exceed the trained distributions (i.e., $\mathcal{N}(z_0; \sqrt{1-\beta_{t}}z_0, \beta_t^2\mathbf{I})$) and its definitive ranges. 

According to Eq.\eqref{misalignment_prevention}, to prevent such misalignment, it is optimal to use a noise schedule that has an explicitly smaller variance during sampling, i.e., \(\forall t \in \lbrack 0,T\rbrack,\ \ \beta_{rev}(t) < \beta_{train}(t)\ \), and therefore the vanilla setting ($\beta_{rev} \equiv \beta_{train}$) in DDIM \cite{DBLP:conf/iclr/SongME21} is sub-optimal in terms of aligning training and sampling. From this perspective, we reveal that noise factor \cite{gao2022difformer} directly benefits from smaller $\beta$ in sampling, and asymmetric time intervals is equivalent to taking a smaller $\beta$: $\beta_{t-\Delta}$ than $\beta_t$ during sampling, thus we show that they are complementary in terms of misalignment prevention during sampling.

\subsection{Connection Between the Two Limitations} For a unified comprehension, we make the following concluding remarks on the connections between the two limitations above. Recall the reverse diffusion process when we first call $f_{\theta}\left( z_{t},x,t \right)$ to obtain a initial $\hat{z}_0$ prediction from pure Gaussian, then call $f_{\theta}\left( z_{t},\hat{z}_0,x,t \right)$ along the diffusion trajectory. When $f_{\theta}$ is \textit{degraded} during training, its denoising capability is thereby hindered, resulting in sub-par prediction accuracy of $\hat{z}_0$ and a greater reconstruction error $\sigma$. The progressive accumulation of $\sigma$ along the trajectory results in an \textit{exacerbated misalignment}, hampering the performance of diffusion.

\begin{figure*}[!tb]
  \centering
  \includegraphics[width=6.5in]{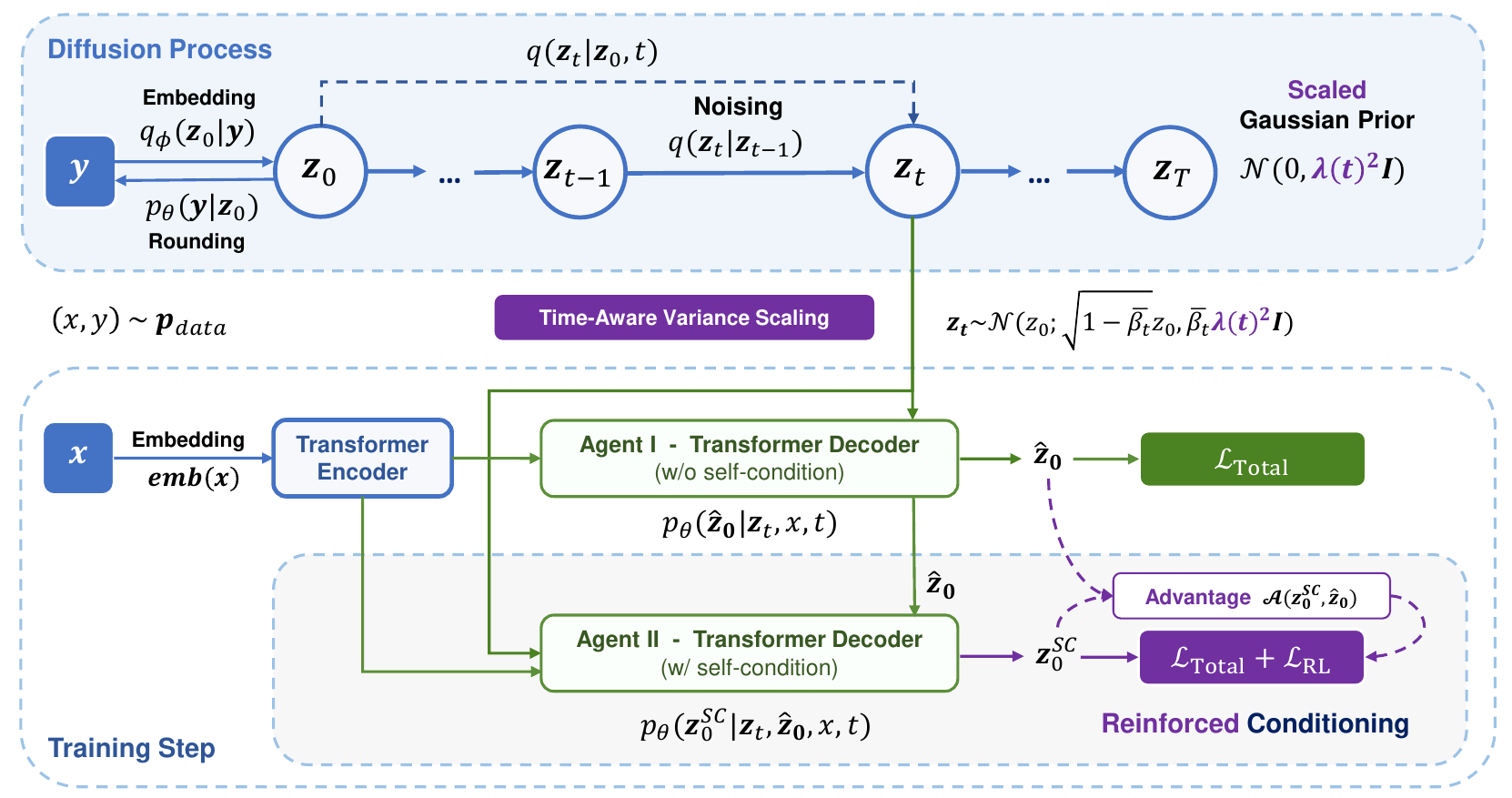}
  \caption{Illustration of \ours, including Reinforced Conditioning and Time-Aware Variance Scaling. }
  \label{methods}
\end{figure*}

\section{Methods}
Our primary motivation is to empower text diffusion \textit{as a whole} through mitigating degradation and misalignment. For the former, we design \textbf{Reinforced Conditioning}, leveraging reinforcement signals to reward quality improvements and enforce the model to better utilize information within noised latent. For the latter, we propose \textbf{Time-Aware Variance Scaling} by increasing model's robustness through accommodating potential sampling errors during training.

Combining the above, we propose \ourswb, namely \textbf{\underline{T}}ext Diffusion model with \textbf{\underline{Re}}inforced \textbf{\underline{C}}onditioning. The design of \ourswb is illustrated in Figure~\ref{methods}. For a pair $(x,y)$ during training, we encode $x$ with a transformer encoder, and $y$ to $z_t$ with word embedding and forward diffusion process. Afterwards, we calculate the advantage from self-conditioning, then back-propagate through the RL objective. Meanwhile, the variance in the forward diffusion process is scaled with a time-aware factor to ensure $ \forall t \in \lbrack 0,T\rbrack, \beta_{rev}(t) < \beta_{train}(t)$. 

\subsection{Reinforced Conditioning}
In this subsection, we provide a RL-based solution to mitigate the degradation of self-conditioning during training. 


\paragraph{Environment and Agents}
We define the environment as conditioned sequence generation task (i.e., $p(y|x)$), with forward diffusion process. For each training step, we first sample data pair $(x,y) \sim p_{data}$ and diffusion time step $t \sim U(0,T)$,  
then embed $x$ via transformer encoder and $y$ via forward diffusion (i.e., $z_0 \leftarrow q_\phi(z_0 |y)), z_t \leftarrow q(z_t|z_0,t)$), 
as illustrated in Figure \ref{methods}. 
We then employ the decoder with and without self-conditioning as two separate agents, namely SC and non-SC agent respectively. Note that they share a same set of parameters $\theta$ (as transformer decoder), but own a different set of policy given their diverse in input conditions. Policies for the non-SC and SC agent could be formalized as:
\begin{equation} 
\begin{split} \pi_\theta^{v} := & \mathop{\arg\max}\limits_{\hat{z}_0} p_\theta(\hat{z}_0 | z_t, x,t); \\  
\pi_\theta^{SC} := & \mathop{\arg\max}\limits_{z_0^{SC}} p_\theta(z_0^{SC} | z_t, \widehat{z}_0, x,t), 
\end{split} 
\end{equation}
with which each agent takes actions to predict starting latent $z_0$ with their input conditions.

\begin{table*}[!t]
\centering
\resizebox{0.8\linewidth}{!}{
\begin{tabular}{l|cc|c|c}
\toprule
\multirow{2}{*}{\textbf{Methods}} & \multicolumn{2}{c}{\begin{tabular}[c]{@{}c@{}}Machine\\ Translation\end{tabular}}    & \multicolumn{1}{|c}{Paraphrase} & \multicolumn{1}{|c}{\begin{tabular}[c]{@{}c@{}}Question\\ Generation\end{tabular}}  \\ \cmidrule(l){2-5} 
& \begin{tabular}[c]{@{}c@{}}\textbf{IWSLT14} \\ De-En\end{tabular} & \begin{tabular}[c]{@{}c@{}}\textbf{WMT14} \\ En-De\end{tabular} 
& \begin{tabular}[c]{@{}c@{}}\textbf{QQP} \end{tabular} & \begin{tabular}[c]{@{}c@{}}\textbf{Quasar-T} \end{tabular} \\ \midrule
Transformer (\citet{vaswani2017attention}) ($b=1$)   & 32.76*  & 26.37* & 30.14\textsuperscript{$\ddag$} & 16.73\textsuperscript{$\ddag$}  \\
Transformer (\citet{vaswani2017attention}) ($b=5$)   & 33.59*  & 27.37*  & 30.86\textsuperscript{$\ddag$} & 17.45\textsuperscript{$\ddag$} \\ 
GPVAE-Finetuned T5 (\citet{du2022diverse}) & - & - & 24.09\textsuperscript{$\dagger$} & 12.51\textsuperscript{$\dagger$} \\\midrule
Levenshetein (\citet{gu2019levenshtein}) ($b=1$)   & -   & 27.27  & 22.68\textsuperscript{$\dagger$} & 9.30\textsuperscript{$\dagger$}    \\
CMLM (\citet{ghazvininejad2019mask}) ($b=10$)  & 33.08   & 27.03  & 24.90 & 7.69     \\
DiffuSeq (\citet{gong2022diffuseq}) ($b=10$)    & 28.78*  & 15.37*  & 24.13 & 17.31    \\
SeqDiffuSeq (\citet{yuan2022seqdiffuseq})  ($b=10$)   & 30.03*  & 17.14* & 24.32 & 17.54  \\ 
DiNoiSer (\citet{ye2023dinoiser}) ($b=5|b=10$)   & 32.23  & 26.08  & 26.07 & -         \\
DiNoiSer (\citet{ye2023dinoiser}) ($b=50|b=20$)  & 32.25  & 26.29  & 25.42 & -         \\
Difformer (\citet{gao2022difformer}) ($b=5 | b=10$) \textsuperscript{$\ddag$}   & 32.01  & 26.89 & 30.58 & 19.55       \\
Difformer (\citet{gao2022difformer}) ($b=20$) \textsuperscript{$\ddag$}  & 32.80  & 27.23 & 30.82 & 20.11        \\ \midrule
\ourswb ($b=5 | b=10$)            & 32.55     & 27.05 & 33.19 & 21.19 \\
\ourswb ($b=20$)           & \textbf{33.31}   & \textbf{27.55} & \textbf{33.26} & \textbf{21.37}           \\ \bottomrule
\end{tabular}}
\vspace{9px}
\caption{BLEU results on sequence generation tasks. `$b$' denotes the beam size for AR Transformer, and the total number of samples used in candidate selection (reranking) for NAR and Diffusion models. $(b=u|b=v)$ denotes a beam size of $u$ and $v$ for the first and last two tasks. We highlight BLEU of the best Non-AR methods in bold. * and $\dagger$ indicates baseline scores quoted from \citet{gao2022difformer} and \citet{gong2022diffuseq}, respectively. $\ddag$ refers to results from our own implementations and experiments.} \label{big2}
\end{table*}

\paragraph{Reward and Training Objective} In designing the training objective, we start from a clear motivation - to tackle the degeneration of self-conditioning by directly rewarding quality improvements and penalizing degrades. To achieve this, we first evaluate the quality of actions: $\widehat{z}_0$ from the non-SC agent $\pi_\theta^{v}$, and $z_0^{SC}$ from the self-conditioning agent $\pi_\theta^{SC}$ with a tractable evaluation metric (i.e., BLEU). Then, we could estimate the advantage of SC agent over non-SC agent as 
\begin{equation}
    \mathcal{A}(z_0^{SC},\widehat{z}_0) = \text{clip}(R(z_0^{SC}) - R(\widehat{z}_0), -\epsilon, \epsilon).
\end{equation}
Inspired by \cite{schulman2017proximal}, we clip the estimated advantages w.r.t. a clipping threshold $\epsilon$ to improve training stability of diffusion. The goal of \ourswb training is to minimize the negative expected advantage:
\begin{equation}
    \label{rlopt}
    \begin{split}
    &\mathcal{L}_{RL}(\theta) = \\
    &-\mathbb{E}_{(x,y) \sim p_{d}, t \sim U(0,T), z_0^{SC} \sim \pi_\theta^{SC}, \widehat{z}_0 \sim \pi_\theta^{v}}\left[ \mathcal{A}(z_0^{SC},\widehat{z}_0)\right].
    \end{split}
\end{equation}
Leveraging REINFORCE \cite{williams1992simple} algorithm, we could thus compute the gradient estimations of Eq.\eqref{rlopt} using a batch of Monte-Carlo samples as follows:
\begin{equation}
    \label{rlloss}
    \nabla_\theta \mathcal{L}_{RL}(\theta) \approx - \frac{1}{N} \sum_{i=1}^{N} \mathcal{A}_i(z_0^{SC},\widehat{z}_0) \nabla_\theta \log p_\theta(y|z_0^{SC}).
\end{equation}
During training, following \citet{chen2022analog}, we take a $50\%$ rate for self-conditioned training. For training steps w/o self-conditioning, we take Eq.\eqref{loss} as training objective, and plug in $\mathcal{L}_{RL}$ (Eq.\eqref{rlloss}) when training w/ self-conditioning. By plugging $\mathcal{L}_{RL}$ into our total objective, we directly mitigate the degradation by providing a clear motivation and guidance on quality gains of $z_0^{SC}$ over $\widehat{z}_0$, thus preventing it from regression to simple repetition of $\widehat{z}_0$ caused by the gradual increasing of $\widehat{z}_0$'s quality during training.

\subsection{Time-Aware Variance Scaling}
\paragraph{Time-Aware Variance Scaling} To alleviate the misalignment brought by the accumulation of prediction error during sampling, we propose an simple but effective method, namely time-aware variance scaling. Specifically, we scale the variance in the forward diffusion process to ensure $ \beta_{rev}(t) < \beta_{train}(t)$, with a \textbf{time-aware} factor $\lambda(t) = k_1+k_2t$, where $k_1,k_2$ denotes hyperparameters of scaling factor. Then, each forward diffusion steps could be expressed as:
\begin{equation}
    q\left({z}_t|{z}_{0}, t\right)=\mathcal{N}\left(z_t ; \sqrt{1-\bar{\beta}_t}z_{0}, \bar{\beta}_t \lambda(t)^2 \mathbf{I}\right) . 
\end{equation}
By scaling variance with $\lambda(t)$, we alter the Gaussian prior for different steps to $\mathcal{N}(0, \lambda(t)^2\mathbf{I})$ during training, while we still sample with the original prior $\mathcal{N}(0, \mathbf{I})$ during sampling. By enlarging the variance scale at each diffusion steps during training, we could increase our method's robustness to the scale of prediction error $\sigma_t$. Since we adapt an increased variance scale during training, we could thus improve model's generation capability by preventing misalignment during sampling. 
Time-dependent scaling is designed to address inconsistent difficulty of diffusion time-steps - denoising a lower noise latent is obviously easier, while reconstruction from higher noise scales (i.e., bigger $t$) is more challenging. 
With time-dependent scaling, we aim to improve further the robustness of preventing misalignment at higher noise scales. In other words, we prevent model from spending excessive effort on training `trivial' low-noise steps, thus facilitating the sufficiency of training.

\begin{table*}[!t]
\centering
\resizebox{0.7\linewidth}{!}{
\begin{tabular}{c|cc|cc|cc}
\toprule
\textbf{Variants} & \begin{tabular}[c]{@{}c@{}}Reinforced \\ Conditioning \end{tabular} 
& \begin{tabular}[c]{@{}c@{}}Variance \\ Scaling \end{tabular} & \begin{tabular}[c]{@{}c@{}} MBR$_{\text{P}}$ \\ ($b=10$) \end{tabular} 
& \begin{tabular}[c]{@{}c@{}} MBR$_{\text{P}}$ \\ ($b=20$) \end{tabular} &
 \begin{tabular}[c]{@{}c@{}} MBR$_{\text{B}}$ \\ ($b=10$) \end{tabular} 
& \begin{tabular}[c]{@{}c@{}} MBR$_{\text{B}}$ \\ ($b=20$) \end{tabular} \\ \midrule
\ourswb (1) & \checkmark & Time-Aware & 33.19 & \textbf{33.26} & 32.11 & 32.60 \\
(2) & $\times$ & Time-Aware & 31.95 & 32.54 & 31.86 & 32.30 \\
(3) & $\times$ & Fixed & 30.48 & 30.71 & 29.90 & 30.49  \\
(4) & $\times$ & $\times$ & 28.08 & 28.85 & 27.31 & 28.49 \\ \bottomrule 
\end{tabular}
}
\vspace{4px}
\caption{Ablation of proposed modules; Comparison of MBR re-ranking metric and candidate set sizes $b$ on the paraphrase (QQP) task. MBR$_{\text{P}}$ and MBR$_{\text{B}}$ denote re-ranking with perplexity or BLEU, respectively. All results reported are BLEU scores.} \label{ablation}
\end{table*}

\section{Experiments}
\subsection{Experiment Setup}
\paragraph{Datasets}
We validate the performance of \ourswb on three important tasks of natural language generation (NLG). Specifically, we select tasks mainly following previous works \cite{gunon, ghazvininejad2019mask, gong2022diffuseq}, including IWSLT14 De-En \cite{cettolo2014report} and WMT14 En-De \cite{bojar2014findings} for translation\footnote{We apply Transformer-Large as teacher for knowledge distillation training in experiments.}, Quasar-T \cite{dhingra2017quasar} for question generation, and Quora (QQP) \cite{chen2018quora} for paraphrase. 

\paragraph{Baselines} We compare our proposed \ourswb with a variety of strong autoregressive, non-autoregressive and diffusion baselines. Specifically, we choose Transformer \cite{vaswani2017attention} and GPVAE \cite{du2022diverse}-Finetuned T5 \cite{raffel2020exploring} for AR models, Levenshtein \cite{gu2019levenshtein}, CMLM \cite{ghazvininejad2019mask} for NAR models, and for diffusion-based models we compare DiffuSeq \cite{gong2022diffuseq}, SeqDiffuSeq \cite{yuan2022seqdiffuseq}, Difformer \cite{gao2022difformer}, including a latest work DiNoiSer \cite{ye2023dinoiser}. 

\paragraph{Implementation}
We adapt \texttt{transformer-base} \cite{vaswani2017attention} architecture of \ourswb ($n_{\text{layers}}=12$, $d_{\text{model}}=512$, $n_{\text{heads}}=8$, $d_{\text{FFN}}=2048$), and set embedding dimension $d=128$. For IWSLT, we reduce $n_{\text{heads}}$ and $d_{\text{FFN}}$ to $4$ and $1024$.
We take $2000$ diffusion steps during training, $20$ during sampling, and apply a $sqrt$ schedule \cite{li2022diffusion}. For time-aware variance scaling, we pick $k_1=3$ and $k_2=7.5e$-$4$ based on preliminary experiments. We train our model on 4 V100 GPUs. We tokenize MT data using \texttt{moses} \cite{artetxeunsupervised}, and learn Byte-Pair Encoding (BPE) \cite{sennrich2016neural}. Following recent advances, we adopt length prediction \cite{lee2020deterministic}, asymmetric decoding \cite{chen2022analog} and MBR decoding \cite{kumar2004minimum} for candidate selection. We apply a learning rate of $5e$-$4$ ($2e$-$4$ for Quasar-T), $10$K warmup steps ($30$K for Quasar-T), and apply the Adam \cite{kingma2014adam} optimizer. 

\subsection{Overall Results}
The experimental results of \ourswb on natural language generation tasks are shown in Table \ref{big2}. As demonstrated in the table, \ourswb surpasses all non-autoregressive and diffusion baselines on a varity of sequence generation tasks (including machine translation, paraphrase and question generation), and also achieves better performance than the autoregressive Transformer on WMT14, QQP and Quasar-T datasets.

Knowledge distillation (KD) is an useful approach in the world of NAR models, and thus we explore \ourswb on machine translation tasks both w/ and w/o KD. As shown in Table \ref{big2}, existing continous diffusion language models, including latest works DiNoiSer \cite{ye2023dinoiser}, fall behind well-established strong NAR baselines, CMLM \cite{savinovstep} for instance. While \ourswb demonstrates strong competitiveness in translation, surpassing CMLM on both datasets. On WMT, \ourswb shows good scalability to larger datasets and greater affinity to knowledge distillation, by being competitive against models in the worlds of NAR and diffusion, as well as the AR Transformer.

\ourswb also show its generic capability on conditioned generation by performing promisingly in question generation (Quasar-T) and paraphrase (QQP). On these tasks, previous Non-AR models fall behind the AR Transformer by a large margin, while \ourswb outperforms AR model significantly. These results demonstrate \ours's strong capability in generating high-quality responses with regard to input contexts. 

\subsection{Detailed Analysis}
In this subsection, we study the effects of the two key parts: Reinforced Conditioning and Time-Aware Variance Scaling.

\begin{figure*}[!t]
  \centerline{
  \includegraphics[width=7.5in]{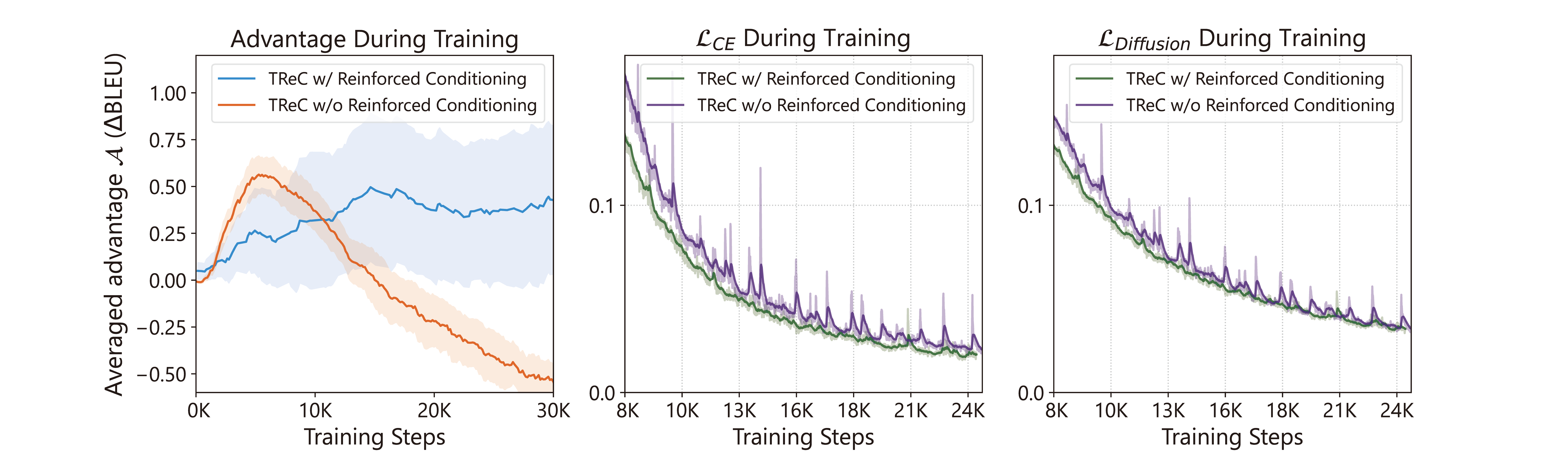}}
  \caption{Degradation Tendency and Training Dynamics for \ourswb w/ and w/o RL on the Quasar task with 3 different seeds.}\label{rl}
\end{figure*}

\paragraph{Mitigation of Degradation} 
To study the effect of $\mathcal{L}_{RL}$ by directly examining the degradation trend of self-conditioning, we use the identical evaluation methods demonstrated in Eq.(\ref{adv_estim}), i.e., evaluating quality advantages of self-conditioning by BLEU metric during training process. As illustrated in Figure \ref{rl}, when training w/o reinforcements, the advantage $\mathcal{A}$ of SC agent over non-SC agent ($\Delta$ BLEU) first rises than drops beyond zero, indicating the degradation of self-conditioning to take place. On the contrary, by adding guidance from $\mathcal{L}_{RL}$ during training, such quality gains from self-conditioning are maintained throughout training, indicating that the trend of degradation is mitigated by reinforcement guidance.

\paragraph{Training Dynamics} 
Moreover, we study the effect of $\mathcal{L}_{RL}$ from another perspective. In design, we plug $\mathcal{L}_{RL}$ into our total objective to mitigate the degradation by providing a clear motivation and guidance on quality gains of $z_0^{SC}$ over $\widehat{z}_0$. To validate the effectiveness of such design, we start from examining the training dynamics of w/ and w/o $\mathcal{L}_{RL}$. As illustrated in Figure \ref{rl}, utilizing reinforced conditioning brings lower losses in both diffusion and cross-entropy part of total loss (Eq.\eqref{loss}) as training progresses, indicating that the $\mathcal{L}_{RL}$ indeed provided helpful guidance for training. Plus, we could also observe that by adding $\mathcal{L}_{RL}$, model enjoys less variance and fluctuations in both part of losses (diffusion and cross-entropy), demonstrating that efforts in preventing degradation do facilitate a stabler training process.

\paragraph{Ablation Study} We study the effect of our proposed modules on model performance in Table \ref{ablation}. We first remove the reinforcement learning module (2), and BLEU scores drop consistently across all sampling settings. We further remove the time-aware variance scaling module (4), and the performance decreased significantly. To test the advantage of our time-aware scaling setting, we replace it with a fixed ratio by removing the time-aware part, and its performance (3) is inferior than time-aware scaling (2). Furthermore, we study effect of various sampling hyper-parameters (MBR re-ranking metric and sampling sizes $b$). As shown in Table \ref{ablation}, Perplexity (via a Transformer with equivalent architecture) outperforms BLEU in re-ranking. Additionally, we observed consistent improvements by increasing candidate sizes $b$, showing model's flexibility to trade-off between cost and quality.

\paragraph{Case Study} 
We present illustrative examples on diffusion process of \ours. These cases demonstrate that \ourswb can generate reasonable sequences through diffusion process. The generation process reveals: (1) \ourswb could quickly generate a high-quality sentence, and converge with only a few steps of iteration (Case \#1). (2) \ourswb is capable of leveraging diffusion process to iteratively refine erroneous predictions for more challenging samples (Case \#2). 

\begin{table}[!t]
\center 
\footnotesize
\resizebox{\linewidth}{!}{
\begin{tabular}{cp{0.79\linewidth}}
\toprule

\textbf{Steps} & \textbf{Input / Reference / Generated Sequence}     \\  \cmidrule(lr){1-2}

\textbf{Input} & does long distance relationship works? \\ 
\textbf{Ref.} & how do i survive in a long distance relationship? \\ \cmidrule(lr){1-2}
1 & how do i work a long distance relationship? \\
2 & how do i work with a long distance relationship? \\
3 & how do i cope with a long distance relationship? \\  \cmidrule(lr){1-2}

\textbf{Input} & if hillary clinton could not continue her presidential campaign, how would the democratic party choose a new candidate? \\
\textbf{Ref.} & if hillary clinton can no longer serve as the democratic nominee, how would her successor be chosen?  \\ \cmidrule(lr){1-2}
1 & if hillary clinton clinton hillary clinton the the the the democratic candidate a presidential candidate?  \\
4 & how hillary clinton to the blue{presidential campaign}, the democratic party choose a presidential candidate? \\
8 & if hillary clinton fails the election, how would the democratic party choose a new candidate? \\ \bottomrule
\end{tabular}%
}
\vspace{10px}
\caption{Cases on QQP. Special tokens are omitted.} \label{cases}
\end{table}

\section{Related Work}
Initial researches in text diffusion models focus on remedy the discreteness of language and adapt diffusion models herein. On this front, \citet{austin2021structured} and \citet{hoogeboom2021argmax} design discrete diffusion based on categorical distributions, while \citet{he2022diffusionbert} explores building diffusion upon state absorption (i.e., masking tokens as noise injection). 
\citet{li2022diffusion} first proposes to directly handle the discreteness by mapping words onto a continuous embedding space. However, the above studies only achieve unconditional or coarse-grained control on generation, whose practical applications are limited.

Consequently, subsequent works mainly focus on conditional generation, which is more practical in NLG. Improvements in the conditioning strategies are mainly categorized three-fold. The first line includes conditioning on controlling attributes, like topics or sentiments \cite{li2023diffusion}, such as \citet{lovelace2022latent} apply class embedding as conditions, \citet{liu2022composable} explore classifier guidance on the latent semantic space for styling controls. The second line applies diffusion models to text-to-text generation, i.e., conditioning on input sequences. This yields more applicable tasks like machine translation or paraphrasing, which are more challenging than conditioning on attributes \cite{li2023diffusion}. For instance, \citet{gao2022difformer} propose partially noising - feeding un-noised conditioning sequences as a reference, while \citet{gong2022diffuseq, gao2022difformer, ye2023dinoiser} encode text condition with an encoder. The third line study conditioning on predictions of previous steps, namely self-conditioning \cite{chen2022analog, strudel2022self} to improve model performance.

Aside from conditioning strategies, other aspects that facilitates text diffusion have also been explored, including balancing embedding scale \cite{yuan2022seqdiffuseq, gao2022difformer}, improving sampling methods \cite{chen2022analog, ye2023dinoiser} and utilizing pretraining \cite{he2022diffusionbert, lin2022genie}. Unlike the existing works, this paper explores a novel conditioning method - reinforced conditioning, which utilizes reward signal to mitigate degradation effect in training text diffusion models. Plus, we propose time-aware variance scaling to better align training and sampling, through alleviating misalignment issue during sampling.

\section{Conclusion}
In this work, we thoroughly analyze the limitations of text diffusion models: degradation during training and misalignment with training during sampling, and propose \ourswb to empower text diffusion with reinforced conditioning and time-aware variance scaling. Our comprehensive experiments demonstrate the competitiveness of \ourswb on multiple language generation tasks, and provide valuable insights into improving training strategies for better diffusion models.

\bibliography{main}

\begin{thebibliography}{43}
\providecommand{\natexlab}[1]{#1}

\bibitem[{Artetxe et~al.(2018)Artetxe, Labaka, Agirre, and Cho}]{artetxeunsupervised}
Artetxe, M.; Labaka, G.; Agirre, E.; and Cho, K. 2018.
\newblock Unsupervised Neural Machine Translation.
\newblock In \emph{International Conference on Learning Representations}.

\bibitem[{Austin et~al.(2021{\natexlab{a}})Austin, Johnson, Ho, Tarlow, and van~den Berg}]{austin2021structureddiff}
Austin, J.; Johnson, D.~D.; Ho, J.; Tarlow, D.; and van~den Berg, R. 2021{\natexlab{a}}.
\newblock Structured denoising diffusion models in discrete state-spaces.
\newblock \emph{Advances in Neural Information Processing Systems}, 34: 17981--17993.

\bibitem[{Austin et~al.(2021{\natexlab{b}})Austin, Johnson, Ho, Tarlow, and van~den Berg}]{austin2021structured}
Austin, J.; Johnson, D.~D.; Ho, J.; Tarlow, D.; and van~den Berg, R. 2021{\natexlab{b}}.
\newblock Structured denoising diffusion models in discrete state-spaces.
\newblock \emph{Advances in Neural Information Processing Systems}, 34: 17981--17993.

\bibitem[{Bojar et~al.(2014)Bojar, Buck, Federmann, Haddow, Koehn, Leveling, Monz, Pecina, Post, Saint-Amand et~al.}]{bojar2014findings}
Bojar, O.; Buck, C.; Federmann, C.; Haddow, B.; Koehn, P.; Leveling, J.; Monz, C.; Pecina, P.; Post, M.; Saint-Amand, H.; et~al. 2014.
\newblock Findings of the 2014 workshop on statistical machine translation.
\newblock In \emph{Proceedings of the ninth workshop on statistical machine translation}, 12--58.

\bibitem[{Cettolo et~al.(2014)Cettolo, Niehues, St{\"u}ker, Bentivogli, and Federico}]{cettolo2014report}
Cettolo, M.; Niehues, J.; St{\"u}ker, S.; Bentivogli, L.; and Federico, M. 2014.
\newblock Report on the 11th IWSLT evaluation campaign.
\newblock In \emph{Proceedings of the 11th International Workshop on Spoken Language Translation: Evaluation Campaign}, 2--17.

\bibitem[{Chen, Zhang, and Hinton(2023)}]{chen2022analog}
Chen, T.; Zhang, R.; and Hinton, G. 2023.
\newblock Analog bits: Generating discrete data using diffusion models with self-conditioning.
\newblock In \emph{International Conference on Learning Representations}.

\bibitem[{Chen et~al.(2018)Chen, Zhang, Zhang, and Zhao}]{chen2018quora}
Chen, Z.; Zhang, H.; Zhang, X.; and Zhao, L. 2018.
\newblock Quora question pairs.

\bibitem[{Dhingra, Mazaitis, and Cohen(2017)}]{dhingra2017quasar}
Dhingra, B.; Mazaitis, K.; and Cohen, W.~W. 2017.
\newblock Quasar: Datasets for question answering by search and reading.
\newblock \emph{arXiv preprint arXiv:1707.03904}.

\bibitem[{Dieleman et~al.(2022)Dieleman, Sartran, Roshannai, Savinov, Ganin, Richemond, Doucet, Strudel, Dyer, Durkan et~al.}]{dieleman2022continuous}
Dieleman, S.; Sartran, L.; Roshannai, A.; Savinov, N.; Ganin, Y.; Richemond, P.~H.; Doucet, A.; Strudel, R.; Dyer, C.; Durkan, C.; et~al. 2022.
\newblock Continuous diffusion for categorical data.
\newblock \emph{arXiv preprint arXiv:2211.15089}.

\bibitem[{Du et~al.(2022)Du, Zhao, Wang, and Ji}]{du2022diverse}
Du, W.; Zhao, J.; Wang, L.; and Ji, Y. 2022.
\newblock Diverse text generation via variational encoder-decoder models with gaussian process priors.
\newblock \emph{arXiv preprint arXiv:2204.01227}.

\bibitem[{Gao et~al.(2022)Gao, Guo, Tan, Zhu, Zhang, Bian, and Xu}]{gao2022difformer}
Gao, Z.; Guo, J.; Tan, X.; Zhu, Y.; Zhang, F.; Bian, J.; and Xu, L. 2022.
\newblock Difformer: Empowering Diffusion Model on Embedding Space for Text Generation.
\newblock \emph{arXiv preprint arXiv:2212.09412}.

\bibitem[{Ghazvininejad et~al.(2019)Ghazvininejad, Levy, Liu, and Zettlemoyer}]{ghazvininejad2019mask}
Ghazvininejad, M.; Levy, O.; Liu, Y.; and Zettlemoyer, L. 2019.
\newblock Mask-Predict: Parallel Decoding of Conditional Masked Language Models.
\newblock In \emph{Proceedings of the 2019 Conference on Empirical Methods in Natural Language Processing and the 9th International Joint Conference on Natural Language Processing}, 6112--6121.

\bibitem[{Gong et~al.(2023)Gong, Li, Feng, Wu, and Kong}]{gong2022diffuseq}
Gong, S.; Li, M.; Feng, J.; Wu, Z.; and Kong, L. 2023.
\newblock {DiffuSeq}: Sequence to Sequence Text Generation with Diffusion Models.
\newblock In \emph{International Conference on Learning Representations}.

\bibitem[{Gu et~al.(2018)Gu, Bradbury, Xiong, Li, and Socher}]{gunon}
Gu, J.; Bradbury, J.; Xiong, C.; Li, V.~O.; and Socher, R. 2018.
\newblock Non-Autoregressive Neural Machine Translation.
\newblock In \emph{International Conference on Learning Representations}.

\bibitem[{Gu, Wang, and Zhao(2019)}]{gu2019levenshtein}
Gu, J.; Wang, C.; and Zhao, J. 2019.
\newblock Levenshtein transformer.
\newblock \emph{Advances in Neural Information Processing Systems}, 32.

\bibitem[{He et~al.(2022)He, Sun, Wang, Huang, and Qiu}]{he2022diffusionbert}
He, Z.; Sun, T.; Wang, K.; Huang, X.; and Qiu, X. 2022.
\newblock DiffusionBERT: Improving Generative Masked Language Models with Diffusion Models.
\newblock \emph{arXiv preprint arXiv:2211.15029}.

\bibitem[{Ho et~al.(2022)Ho, Chan, Saharia, Whang, Gao, Gritsenko, Kingma, Poole, Norouzi, Fleet et~al.}]{ho2022imagen}
Ho, J.; Chan, W.; Saharia, C.; Whang, J.; Gao, R.; Gritsenko, A.; Kingma, D.~P.; Poole, B.; Norouzi, M.; Fleet, D.~J.; et~al. 2022.
\newblock Imagen video: High definition video generation with diffusion models.
\newblock \emph{arXiv preprint arXiv:2210.02303}.

\bibitem[{Ho, Jain, and Abbeel(2020)}]{ho2020denoising}
Ho, J.; Jain, A.; and Abbeel, P. 2020.
\newblock Denoising diffusion probabilistic models.
\newblock \emph{Advances in Neural Information Processing Systems}, 33: 6840--6851.

\bibitem[{Hoogeboom et~al.(2021)Hoogeboom, Nielsen, Jaini, Forr{\'e}, and Welling}]{hoogeboom2021argmax}
Hoogeboom, E.; Nielsen, D.; Jaini, P.; Forr{\'e}, P.; and Welling, M. 2021.
\newblock Argmax flows and multinomial diffusion: Learning categorical distributions.
\newblock \emph{Advances in Neural Information Processing Systems}, 34: 12454--12465.

\bibitem[{Kingma and Ba(2015)}]{kingma2014adam}
Kingma, D.~P.; and Ba, J. 2015.
\newblock Adam: {A} Method for Stochastic Optimization.
\newblock In \emph{3rd International Conference on Learning Representations}.

\bibitem[{Kong et~al.(2021)Kong, Ping, Huang, Zhao, and Catanzaro}]{kongdiffwave}
Kong, Z.; Ping, W.; Huang, J.; Zhao, K.; and Catanzaro, B. 2021.
\newblock DiffWave: A Versatile Diffusion Model for Audio Synthesis.
\newblock In \emph{International Conference on Learning Representations}.

\bibitem[{Kumar and Byrne(2004)}]{kumar2004minimum}
Kumar, S.; and Byrne, W.~J. 2004.
\newblock Minimum Bayes-Risk Decoding for Statistical Machine Translation.
\newblock In \emph{Human Language Technology Conference of the North American Chapter of the Association for Computational Linguistics}, 169--176. The Association for Computational Linguistics.

\bibitem[{Lee, Mansimov, and Cho(2020)}]{lee2020deterministic}
Lee, J.; Mansimov, E.; and Cho, K. 2020.
\newblock Deterministic non-autoregressive neural sequence modeling by iterative refinement.
\newblock In \emph{2018 Conference on Empirical Methods in Natural Language Processing}, 1173--1182. Association for Computational Linguistics.

\bibitem[{Li et~al.(2022)Li, Thickstun, Gulrajani, Liang, and Hashimoto}]{li2022diffusion}
Li, X.; Thickstun, J.; Gulrajani, I.; Liang, P.~S.; and Hashimoto, T.~B. 2022.
\newblock Diffusion-lm improves controllable text generation.
\newblock \emph{Advances in Neural Information Processing Systems}, 35: 4328--4343.

\bibitem[{Li et~al.(2023)Li, Zhou, Zhao, and Wen}]{li2023diffusion}
Li, Y.; Zhou, K.; Zhao, W.~X.; and Wen, J.-R. 2023.
\newblock Diffusion Models for Non-autoregressive Text Generation: A Survey.
\newblock \emph{arXiv preprint arXiv:2303.06574}.

\bibitem[{Lin et~al.(2022)Lin, Gong, Shen, Wu, Fan, Lin, Chen, and Duan}]{lin2022genie}
Lin, Z.; Gong, Y.; Shen, Y.; Wu, T.; Fan, Z.; Lin, C.; Chen, W.; and Duan, N. 2022.
\newblock GENIE: Large Scale Pre-training for Text Generation with Diffusion Model.
\newblock \emph{arXiv preprint arXiv:2212.11685}.

\bibitem[{Liu et~al.(2022{\natexlab{a}})Liu, Feng, Gao, Yang, Liang, Bao, He, Cui, Li, and Hu}]{liu2022composable}
Liu, G.; Feng, Z.; Gao, Y.; Yang, Z.; Liang, X.; Bao, J.; He, X.; Cui, S.; Li, Z.; and Hu, Z. 2022{\natexlab{a}}.
\newblock Composable Text Controls in Latent Space with ODEs.
\newblock \emph{arXiv preprint arXiv:2208.00638}.

\bibitem[{Liu et~al.(2022{\natexlab{b}})Liu, Li, Ren, Chen, and Zhao}]{liu2022diffsinger}
Liu, J.; Li, C.; Ren, Y.; Chen, F.; and Zhao, Z. 2022{\natexlab{b}}.
\newblock Diffsinger: Singing voice synthesis via shallow diffusion mechanism.
\newblock In \emph{Proceedings of the AAAI Conference on Artificial Intelligence}, volume~36, 11020--11028.

\bibitem[{Lovelace et~al.(2022)Lovelace, Kishore, Wan, Shekhtman, and Weinberger}]{lovelace2022latent}
Lovelace, J.; Kishore, V.; Wan, C.; Shekhtman, E.; and Weinberger, K. 2022.
\newblock Latent Diffusion for Language Generation.
\newblock \emph{arXiv preprint arXiv:2212.09462}.

\bibitem[{Raffel et~al.(2020)Raffel, Shazeer, Roberts, Lee, Narang, Matena, Zhou, Li, and Liu}]{raffel2020exploring}
Raffel, C.; Shazeer, N.; Roberts, A.; Lee, K.; Narang, S.; Matena, M.; Zhou, Y.; Li, W.; and Liu, P.~J. 2020.
\newblock Exploring the limits of transfer learning with a unified text-to-text transformer.
\newblock \emph{The Journal of Machine Learning Research}, 21(1): 5485--5551.

\bibitem[{Rombach et~al.(2022)Rombach, Blattmann, Lorenz, Esser, and Ommer}]{rombach2022high}
Rombach, R.; Blattmann, A.; Lorenz, D.; Esser, P.; and Ommer, B. 2022.
\newblock High-resolution image synthesis with latent diffusion models.
\newblock In \emph{Proceedings of the IEEE/CVF Conference on Computer Vision and Pattern Recognition}, 10684--10695.

\bibitem[{Savinov et~al.(2022)Savinov, Chung, Binkowski, Elsen, and van~den Oord}]{savinovstep}
Savinov, N.; Chung, J.; Binkowski, M.; Elsen, E.; and van~den Oord, A. 2022.
\newblock Step-unrolled Denoising Autoencoders for Text Generation.
\newblock In \emph{International Conference on Learning Representations}.

\bibitem[{Schulman et~al.(2017)Schulman, Wolski, Dhariwal, Radford, and Klimov}]{schulman2017proximal}
Schulman, J.; Wolski, F.; Dhariwal, P.; Radford, A.; and Klimov, O. 2017.
\newblock Proximal policy optimization algorithms.
\newblock \emph{arXiv preprint arXiv:1707.06347}.

\bibitem[{Sennrich, Haddow, and Birch(2016)}]{sennrich2016neural}
Sennrich, R.; Haddow, B.; and Birch, A. 2016.
\newblock Neural Machine Translation of Rare Words with Subword Units.
\newblock In \emph{Proceedings of the 54th Annual Meeting of the Association for Computational Linguistics (Volume 1: Long Papers)}, 1715--1725.

\bibitem[{Sohl-Dickstein et~al.(2015)Sohl-Dickstein, Weiss, Maheswaranathan, and Ganguli}]{sohl2015deep}
Sohl-Dickstein, J.; Weiss, E.; Maheswaranathan, N.; and Ganguli, S. 2015.
\newblock Deep unsupervised learning using nonequilibrium thermodynamics.
\newblock In \emph{International Conference on Machine Learning}, 2256--2265. PMLR.

\bibitem[{Song, Meng, and Ermon(2021)}]{DBLP:conf/iclr/SongME21}
Song, J.; Meng, C.; and Ermon, S. 2021.
\newblock Denoising Diffusion Implicit Models.
\newblock In \emph{9th International Conference on Learning Representations}.

\bibitem[{Strudel et~al.(2022)Strudel, Tallec, Altch{\'e}, Du, Ganin, Mensch, Grathwohl, Savinov, Dieleman, Sifre et~al.}]{strudel2022self}
Strudel, R.; Tallec, C.; Altch{\'e}, F.; Du, Y.; Ganin, Y.; Mensch, A.; Grathwohl, W.; Savinov, N.; Dieleman, S.; Sifre, L.; et~al. 2022.
\newblock Self-conditioned embedding diffusion for text generation.
\newblock \emph{arXiv preprint arXiv:2211.04236}.

\bibitem[{Vahdat, Kreis, and Kautz(2021)}]{vahdat2021score}
Vahdat, A.; Kreis, K.; and Kautz, J. 2021.
\newblock Score-based generative modeling in latent space.
\newblock \emph{Advances in Neural Information Processing Systems}, 34: 11287--11302.

\bibitem[{Vaswani et~al.(2017)Vaswani, Shazeer, Parmar, Uszkoreit, Jones, Gomez, Kaiser, and Polosukhin}]{vaswani2017attention}
Vaswani, A.; Shazeer, N.; Parmar, N.; Uszkoreit, J.; Jones, L.; Gomez, A.~N.; Kaiser, {\L}.; and Polosukhin, I. 2017.
\newblock Attention is all you need.
\newblock \emph{Advances in neural information processing systems}, 30.

\bibitem[{Wehenkel and Louppe(2021)}]{wehenkeldiffusion}
Wehenkel, A.; and Louppe, G. 2021.
\newblock Diffusion Priors In Variational Autoencoders.
\newblock In \emph{ICML Workshop on Invertible Neural Networks, Normalizing Flows, and Explicit Likelihood Models}.

\bibitem[{Williams(1992)}]{williams1992simple}
Williams, R.~J. 1992.
\newblock Simple statistical gradient-following algorithms for connectionist reinforcement learning.
\newblock \emph{Reinforcement learning}, 5--32.

\bibitem[{Ye et~al.(2023)Ye, Zheng, Bao, Qian, and Wang}]{ye2023dinoiser}
Ye, J.; Zheng, Z.; Bao, Y.; Qian, L.; and Wang, M. 2023.
\newblock DINOISER: Diffused Conditional Sequence Learning by Manipulating Noises.
\newblock \emph{arXiv preprint arXiv:2302.10025}.

\bibitem[{Yuan et~al.(2022)Yuan, Yuan, Tan, Huang, and Huang}]{yuan2022seqdiffuseq}
Yuan, H.; Yuan, Z.; Tan, C.; Huang, F.; and Huang, S. 2022.
\newblock SeqDiffuSeq: Text Diffusion with Encoder-Decoder Transformers.
\newblock \emph{arXiv preprint arXiv:2212.10325}.

\end{thebibliography}
\end{document}